\documentclass{article}
\usepackage{ijcai18}

\usepackage{xcolor}
\usepackage{soul}
\usepackage[utf8]{inputenc}
\usepackage[small]{caption}
\usepackage{lineno}
\usepackage[draft]{hyperref}
\usepackage{multirow}
\usepackage{bbm}
\usepackage{bm}
\usepackage{amsmath}
\usepackage{amssymb}
\usepackage{algorithm}
\usepackage[noend]{algorithmic}
\usepackage{booktabs}
\usepackage{array}
\usepackage{graphicx}
\usepackage{libertine}
\usepackage{times}
\title{Adversarially Regularized Graph Autoencoder for Graph Embedding}

\author{
Shirui Pan$^{1}$\thanks{These authors contribute  equally to this work.}, 
Ruiqi Hu$^{1*}$, 
Guodong Long$^1$, 
Jing Jiang$^1$, 
Lina Yao$^2$, 
Chengqi Zhang$^1$
\\ 
$^1$ Centre for Artificial Intelligence, FEIT, University of Technology Sydney, Australia \\
$^2$ School of Computer Science and Engineering, University of New South Wales, Australia  \\
shirui.pan@uts.edu.au,
ruiqi.hu@student.uts.edu.au,
guodong.long@uts.edu.au, \\
jing.jiang@uts.edu.au,
lina.yao@unsw.edu.au,
chengqi.zhang@uts.edu.au
}

\begin{document}

\maketitle

\begin{abstract}

Graph embedding is an effective method to represent graph data in a low dimensional space for graph analytics.  Most existing embedding algorithms typically focus on preserving the topological structure or minimizing the reconstruction errors of graph data,  but they have mostly ignored the data distribution of the latent codes from the graphs, which often results in inferior embedding in  real-world  graph data. In this paper, we propose a novel adversarial graph embedding framework for graph data. The framework encodes the topological structure and node content in a graph to a compact representation, on which a decoder is trained to reconstruct the graph structure. Furthermore, the latent representation is enforced to match a prior distribution via an adversarial training scheme. To learn a robust embedding,  two variants of adversarial approaches,  \textit{adversarially regularized graph autoencoder} (ARGA) and \textit{adversarially regularized variational graph autoencoder} (ARVGA), are developed. Experimental studies on real-world graphs validate our design and demonstrate that our algorithms outperform baselines by a wide margin in link prediction,  graph clustering, and graph visualization tasks.
\end{abstract}

\section{Introduction}
Graphs are essential tools to capture and model  complicated relationships among data. 
 In a variety of graph applications, including protein-protein interaction networks, social media, and citation networks, analyzing  graph data plays an important role in various data mining tasks including node or graph classification \cite{kipf2016semi,pan2016joint}, link prediction \cite{wang2017predictive}, and node clustering \cite{wang2017mgae}. However, the high computational complexity, low parallelizability, and inapplicability of machine learning methods to graph data have made these graph analytic tasks profoundly challenging \cite{cui2017survey}. Recently \textit{graph embedding} has emerged as a general approach to these problems.

Graph embedding converts  graph data into
a low dimensional, compact, and continuous feature space. The key idea is to preserve the topological structure, vertex content, and other side information \cite{zhang2017network}. This new learning paradigm has shifted the tasks of seeking complex models for classification, clustering, and link prediction to learning a robust representation of the graph data, so that  any graph analytic task can be easily performed by employing simple traditional models (e.g.,  a linear SVM for the classification task). This merit has motivated a number of studies in this area \cite{cai2017comprehensive,goyal2017graph}.

Graph embedding algorithms can be classified into three categories:  probabilistic models, matrix factorization-based algorithms, and deep learning-based algorithms. Probabilistic models like DeepWalk \cite{perozzi2014deepwalk}, node2vec \cite{grover2016node2vec} and LINE \cite{Tang2015} attempt to learn graph embedding by extracting different patterns from the graph.  The captured patterns or walks include global structural equivalence, local neighborhood connectivities, and other various order proximities. Compared with classical methods such as Spectral Clustering \cite{tang2011leveraging}, these graph embedding algorithms perform more effectively and are scalable to large graphs.


Matrix factorization-based algorithms, such as GraRep \cite{cao2015grarep}, HOPE \cite{ou2016asymmetric}, M-NMF \cite{wang2017community} pre-process the graph structure into an adjacency matrix and get the embedding by decomposing the adjacency matrix.  Recently it has been shown that many probabilistic algorithms are equivalent to matrix factorization approaches \cite{qiu2017network}. Deep learning approaches, especially autoencoder-based methods,  are also widely studied for graph embedding. SDNE \cite{wang2016structural} and DNGR \cite{cao2016deep} employ deep autoencoders to preserve the graph proximities and model positive pointwise mutual information (PPMI). The MGAE algorithm utilizes a marginalized single layer autoencoder to learn representation for clustering \cite{wang2017mgae}. 

%

The approaches above are typically unregularized approaches which mainly focus on preserving the structure relationship (probabilistic approaches), or minimizing the reconstruction error (matrix factorization or deep learning methods). They have mostly ignored the data distribution of the latent codes. In practice unregularized embedding approaches often learn a degenerate \textit{identity} mapping where the latent code space is free of any structure \cite{makhzani2015adversarial}, and can easily result in poor representation in dealing with real-world sparse and noisy graph data. 
One common way to handle this problem is to introduce some regularization to the latent codes and enforce them to follow some prior data distribution \cite{makhzani2015adversarial}. Recently generative adversarial based frameworks  \cite{donahue2016adversarial,radford2015unsupervised} have also been developed for learning robust latent representation. However, none of these frameworks is specifically for graph data, where both topological structure and content information are required to embed to a latent space.



In this paper, we propose a novel adversarial framework with two variants, namely \textit{adversarially regularized graph autoencoder} (ARGA) and \textit{adversarially regularized variational graph autoencoder} (ARVGA),  for graph embedding. 
The theme of our framework is to not only minimize the reconstruction errors of the graph structure but also to enforce the latent codes to match a prior distribution. By exploiting both graph structure and node content with a graph convolutional network, our algorithms encodes the graph data in the latent space. With a decoder aiming at reconstructing the topological graph information, 
we further incorporate an adversarial training scheme to regularize the latent codes to learn a robust graph representation. The adversarial training module aims to discriminate if the latent codes are from a real prior distribution or from the graph encoder. The graph encoder learning and adversarial regularization are jointly optimized in a unified framework so that each can be beneficial to the other and finally lead to a better graph embedding. The experimental results on benchmark datasets demonstrate the superb performance of our algorithms on three unsupervised graph analytic tasks, namely link prediction, node clustering, and graph visualization. Our contributions can be summarized below:


\begin{itemize}
\item We propose a novel adversarially regularized framework for graph embedding, which represent topological structure and node content in a continuous vector space. Our framework  learns the embedding to minimize the reconstruction error while enforcing the latent codes to match a prior distribution. 

\item We develop two variants of adversarial  approaches, \textit{adversarially regularized graph autoencoder} (ARGA) and \textit{adversarially regularized variational graph autoencoder} (ARVGA) to learn the graph embedding. 

\item  Experiments on  benchmark graph datasets  demonstrate  that our graph embedding approaches outperform the others on three unsupervised tasks.
\end{itemize}

\section{Related Work}

\noindent\textbf{Graph Embedding Models}. 
From the perspective of information exploration, graph embedding algorithms can be also separated into two groups: topological embedding approaches and content enhanced embedding methods. 

Topological embedding approaches assume that there is only topological structure information available, and the learning objective is to preserve the topological information maximumly. 
Perozzi et al. propose a DeepWalk model to learn the node embedding from a collection of random walks \cite{perozzi2014deepwalk}. Since then, a number of probabilistic models such as node2vec \cite{grover2016node2vec} and LINE \cite{Tang2015} have been developed. As a graph can be mathematically represented as an adjacency matrix, many matrix factorization approaches such as GraRep \cite{cao2015grarep}, HOPE \cite{ou2016asymmetric}, M-NMF \cite{wang2017community} are proposed to learn the latent representation for a graph. Recently deep learning models have been widely exploited to learn the graph embedding. These algorithms preserve the first and second order of proximities \cite{wang2016structural}, or reconstruct the positive pointwise mutual information (PPMI) \cite{cao2016deep} via different variants of autoencoders.



Content enhanced embedding methods assume  node content information is available and exploit both topological information and content features simultaneously. TADW \cite{yang2015network} presents a matrix factorization approach to explore node features. TriDNR \cite{pan2016tri} captures structure, node content, and label information via a tri-party neural network architecture. UPP-SNE employs an approximated kernel mapping scheme to  exploit user profile features to enhance the embedding learning of users in social networks \cite{zhang2017user}.

Unfortunately the above algorithms largely ignore the latent distribution of the embedding, which may result in poor representation in practice. In this paper, we explore  adversarial training methods to address this issue.

\noindent\textbf {Adversarial Models.}
Our method is motivated by the generative adversarial network (GAN) \cite{goodfellow2014generative}. GAN plays an adversarial game with two linked models: the generator $\mathcal{G}$ and the discriminator $\mathcal{D}$. The discriminator can be a  multi-layer perceptron which discriminates if an input sample comes from the data distribution or from the generator we built. Simultaneously, the generator is trained to generate the samples to convince the discriminator that the generated samples come from the prior data distribution. 
Due to its effectiveness in many unsupervised tasks, recently a number of adversarial training algorithms have been proposed  \cite{donahue2016adversarial,radford2015unsupervised}.

Recently Makhzani et al. proposed an adversarial autoencoder (AAE) to learn the latent embedding by merging the adversarial mechanism into the autoencoder \cite{makhzani2015adversarial}. However, it is designed for general data rather than graph data. Dai et al. applied the adversarial mechanism to graphs. However, their approach can only exploit the topological information \cite{dai2017adversarial}. In contrast, our algorithm is more flexible and can handle both topological and content information for graph data. 

\begin{figure*}
\centering
\includegraphics[ width=.9\linewidth]{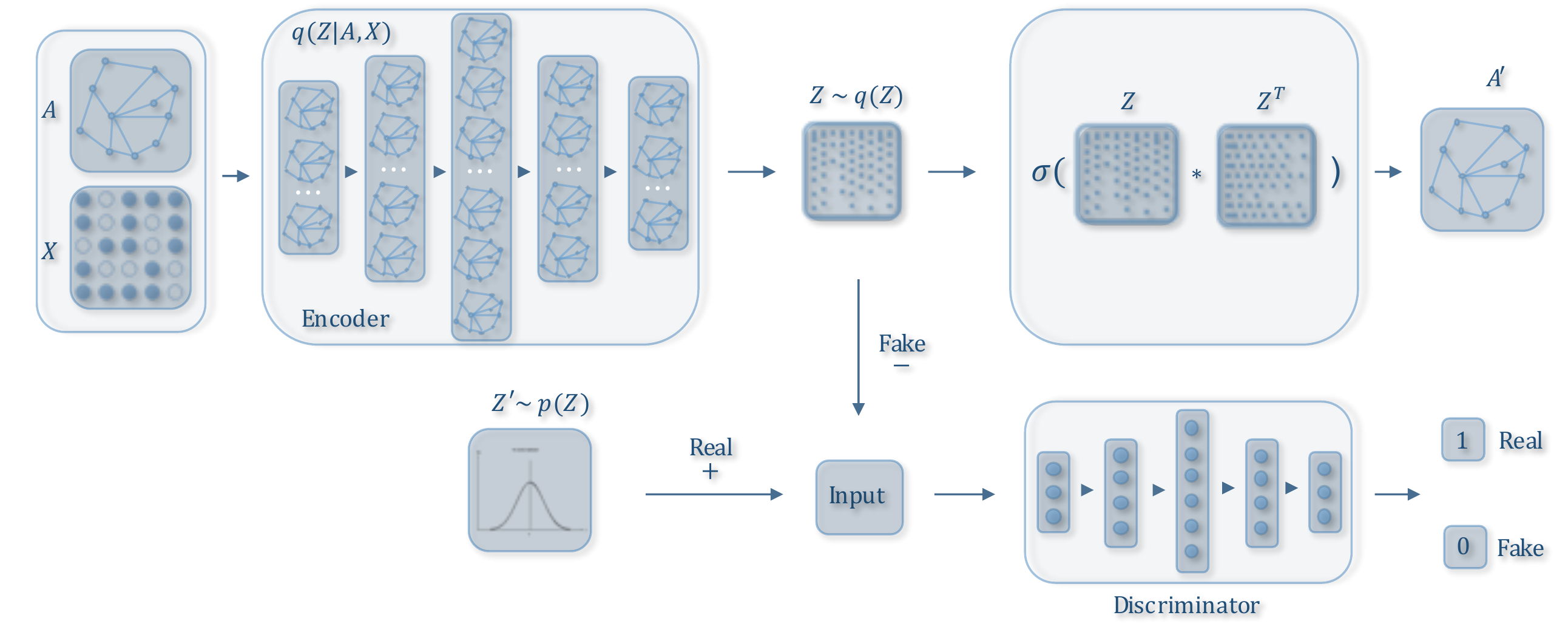}
\caption{The architecture of the adversarially regularized graph autoencoder (ARGA). The upper tier is a graph convolutional autoencoder that reconstructs a graph $\mathbf{A}$ from an embedding  $\mathbf{Z}$ which is generated by the encoder which exploits graph structure $\mathbf{A}$ and the node content matrix $\mathbf{X}$. The lower tier is an adversarial network trained to discriminate if a sample is generated from the embedding or from a prior distribution.  The adversarially regularized variational graph autoencoder (ARVGA) is similar to ARGA except that it employs a \textit{variational} graph autoencoder in the upper tier (See Algorithm 1 for details). 
} \label{fig:construction}
\end{figure*}

\section{Problem Definition and Framework}
A graph is represented as $\mathbf{G} = \{\mathbf{V}, \mathbf{E}, \mathbf{X}\}$, where $\mathbf{V} = \{\mathbf{v}_i\}_i = 1, \cdots, n$ consists of a set of nodes in a graph and $\mathbf{e}_{i,j}=<\mathbf{v}_{i},\mathbf{v}_{j}>  \in \mathbf{E}$ represents a linkage encoding the citation edge between the nodes. The topological structure of graph $\mathbf{G}$ can be represented by an adjacency matrix $\mathbf{A}$, where $\mathbf{A}_{i,j} = 1$ if $\mathbf{e}_{i,j} \in \mathbf{E}$, otherwise $\mathbf{A}_{i,j} = 0$. $\mathbf{x}_i \in \mathbf{X}$ indicates the content features associated with each node $\mathbf{v}_i$. 

Given a graph $\mathbf{G}$, our purpose is to map the nodes $\mathbf{v}_{i} \in \mathbf{V}$ to  low-dimensional vectors $\mathbf{z}_i \in \mathbb{R}^d$ with the formal format as follows: $f: \mathbf{(A, X)} \rightarrowtail \mathbf{Z}$, where $\mathbf{z}_i^{\top}$ is the $i$-th row of the matrix $\mathbf{Z} \in \mathbb{R}^{n \times d}$. $n$ is the number of nodes and $d$ is the dimension of embedding. We take $\mathbf{Z}$ as the embedding matrix and the embeddings should well preserve the topological structure $\mathbf{A}$ as well as content information $\mathbf{X}$ .

\subsection{Overall Framework}
Our objective is to learn a robust embedding given a graph $\mathbf{G} = \{\mathbf{V}, \mathbf{E}, \mathbf{X}\}$. To this end, we leverage an adversarial architecture with a graph autoencoder to directly process the entire graph and learn a robust embedding. Figure \ref{fig:construction} demonstrates the workflow of ARGA which consists of two modules: the graph autoencoder and the adversarial network.
\begin{itemize}
\item \textbf{Graph Convolutional Autoencoder}. The autoencoder takes in the structure of graph $\mathbf{A}$ and the node content $\mathbf{X}$ as inputs to learn a latent representation $\mathbf{Z}$, and then reconstructs the graph structure $\mathbf{A}$ from $\mathbf Z$.
\item \textbf{Adversarial Regularization.} The adversarial network forces the latent codes to match a prior distribution by an adversarial training module, which discriminates whether the current latent code $\mathbf z_i \in \mathbf{Z}$ comes from the encoder or from the prior distribution. 
\end{itemize}

\section{Proposed Algorithm}
\subsection{Graph Convolutional Autoencoder}
The graph convolutional autoencoder aims to embed a graph $\mathbf{G} = \{\mathbf{V}, \mathbf{E}, \mathbf{X}\}$ in a low-dimensional space. Two key questions arise (1) how to integrate both graph structure $\mathbf{A}$ and node content $\mathbf{X}$ in an encoder, and (2)  what sort of information should be reconstructed via a decoder? 

\vspace{.1cm}
 \noindent \textbf{Graph Convolutional Encoder Model  $\mathcal{G}(\mathbf X, \mathbf A)$. \ } 
 To represent both graph structure $\mathbf{A}$ and node content $\mathbf{X}$ in a unified framework, we  develop a variant of the graph convolutional network (GCN) \cite{kipf2016semi} as a graph encoder. Our graph convolutional network (GCN) extends the operation of \textit{convolution} to graph data in the spectral domain, and learns a layer-wise transformation by a spectral convolution function $f(\mathbf{Z}^{(l)},\mathbf{A} |\mathbf{W}^{(l)})$:
 \begin{equation}
\mathbf{Z}^{(l+1)} = f(\mathbf{Z}^{(l)},\mathbf{A} |\mathbf{W}^{(l)})
\label{eq:encoder}
\end{equation}
Here, $\mathbf{Z}^l $  is the input for convolution, and $\mathbf{Z}^{(l+1)}$ is the output after convolution. We  have $\mathbf{Z}^0 = \mathbf{X} \in \mathbb{R}^{n\times m}$  ($n$ nodes and $m$ features) for our problem. $\mathbf{W}^{(l)} $ is a matrix of filter parameters we need to learn in the neural network. If $f(\mathbf{Z}^{(l)},\mathbf{A} |\mathbf{W}^{(l)})$  is well defined, we can build arbitrary deep convolutional neural networks efficiently.

Each layer of our graph convolutional network can be expressed with the function $f(\mathbf{Z}^{(l)},\mathbf{A} |\mathbf{W}^{(l)})$ as follows:
\begin{equation}
\label{eq:convolutional}
f(\mathbf{Z}^{(l)},\mathbf{A} |\mathbf{W}^{(l)})=   \phi(\widetilde{\mathbf{D}}^{-\frac{1}{2}}\widetilde{\mathbf{A}}\widetilde{\mathbf{D}}^{-\frac{1}{2}}\mathbf{Z}^{(l)}\mathbf{W}^{(l)}),
\end{equation}
where $\widetilde{\mathbf{A}} = \mathbf{A}+ \mathbf{I}$ and $\widetilde{\mathbf{D}}_{ii} = \sum_j\widetilde{\mathbf{A}}_{ij}$. $\mathbf{I}$ is the identity matrix of ${\mathbf{A}}$ and $\phi$ is an activation function such as $\text{Relu}(t) = \max(0,t)$ or $\text{sigmoid}(t)=\frac{1}{1+e^t}$. 
Overall, the graph encoder $\mathcal{G}(\mathbf X, \mathbf A)$ is constructed with a two-layer GCN. In our paper, we develop two variants of  encoder, e.g., Graph Encoder and Variational Graph Encoder.

The \noindent\textit{ Graph Encoder} is constructed as follows: 
\begin{eqnarray}
 \mathbf{Z}^{(1)}= f_{\text{Relu}}(\mathbf X,\mathbf{A}|  \mathbf W^{(0)}); \label{eq:z1}\\
 \mathbf Z^{(2)} = f_{\text{linear}}(\mathbf{Z}^{(1)},\mathbf{A} | \mathbf W^{(1)}). \label{eq:latent}
\end{eqnarray}
 $\text{Relu}(\cdot)$ and linear activation  functions are used for the first and second layers. Our graph convolutional encoder  $\mathcal{G}(\mathbf Z, \mathbf A)=q(\mathbf{Z}|\mathbf{X}, \mathbf{A})$  encodes both graph structure and node content into a  representation $\mathbf Z= q(\mathbf{Z}|\mathbf{X}, \mathbf{A})=\mathbf{Z}^{(2)}$.

\vspace{.1cm}
A \noindent\textit{Variational Graph Encoder}  is defined by an inference model:
\begin{eqnarray}
	q(\mathbf{Z}|\mathbf{X}, \mathbf{A}) = \prod^{n}_{i=1}q(\mathbf{z_i}|\mathbf{X},\mathbf{A}),\\
		q(\mathbf{z_i}|\mathbf{X},\mathbf{A}) = \mathcal{N}(\mathbf{z}_i|  \bm{{\mu}}_i, \text{diag}(\bm \sigma^2))
\end{eqnarray}

Here, $ \bm \mu= \mathbf Z^{(2)}$ is the matrix of mean vectors $\bm z_i$
; similarly $\text{log}  \bm \sigma = f_{\text{linear}}(\mathbf{Z}^{(1)},\mathbf{A} | \mathbf W'^{(1)})$ which share the weights $\mathbf W^{(0)}$ with $ \bm \mu$ in the first layer in Eq. (\ref{eq:z1}).

\vspace{.1cm}
\noindent \textbf{Decoder Model. \ } Our decoder model is used to reconstruct the graph data. We can reconstruct either the graph structure $\mathbf A$, content information $\mathbf X$, or both. In our paper, we propose to reconstruct graph structure $\mathbf A$, which  provides more flexibility in the sense that our algorithm will still function properly even if there is no content information $\mathbf X$ available (e.g., $\mathbf X = \mathbf I$).
 Our decoder $p(\hat{\mathbf{A}}|\mathbf{Z})$ predicts whether there is a link between two nodes. More specifically, we train a link prediction layer based on the graph embedding:
 \begin{eqnarray}
 \small
 	p(\hat{\mathbf{A}}|\mathbf{Z}) = \prod^n_{i=1} \prod^n_{j=1} p(\hat{\mathbf{A}}_{ij}| \mathbf{z}_i,\mathbf{z}_j);\\
 		p(\hat{\mathbf{A}}_{ij} = 1 |\mathbf{z}_i, \mathbf{z}_j) = \text{sigmoid}(\mathbf{z}^{\top}_i, \mathbf{z}_j),
 \end{eqnarray}

\vspace{.1cm}
\noindent \textbf{Graph Autoencoder Model. \ } 
The embedding $\mathbf{Z}$ and the reconstructed graph $\hat{\mathbf{A}}$ can be presented as follows:
\begin{equation}
	\hat{\mathbf{A}} =\text{sigmoid}(\mathbf{Z}\mathbf{Z}^{\top}),\ \text{here} \ \mathbf{Z}  = q(\mathbf{Z}|\mathbf{X}, \mathbf{A}) 
\end{equation}

\noindent \textbf{Optimization. \ }  For the graph encoder, we minimize the reconstruction error of the graph data by:
\begin{equation}
	\mathcal{L}_{0} = \mathbb{E}_{q(\mathbf{Z|(X,A)})}[\text{log}~ p(\hat{\mathbf{A}}|\mathbf{Z})] \label{eq:autoencoder_obj0}
\end{equation}
For the variational graph encoder, we optimize the variational lower bound as follows:
\begin{equation}
	\mathcal{L}_{1} = \mathbb{E}_{q(\mathbf{Z|(X,A)})}[\text{log}~ p(\hat{\mathbf{A}}|\mathbf{Z})] - \mathbf{KL}[q(\mathbf{Z|\mathbf{X}},\mathbf{A}) \parallel p(\mathbf{Z})] \label{eq:autoencoder_obj}
\end{equation}
where $\mathbf{KL}[q(\bullet)||p(\bullet)]$ is the Kullback-Leibler divergence between $q(\bullet)$ and $p(\bullet)$. We also take a Gaussian prior $p(\mathbf{Z}) = \prod_i p(\mathbf{z}_i) = \prod_i\mathcal{N}(\mathbf{z}_i|0,\mathbf{I}) $. 

\subsection{Adversarial Model $\mathcal{D}(\mathbf Z)$}
The key idea of our  model is to enforce latent representation $\mathbf Z$ to match a prior distribution, which is achieved by an adversarial training model. The adversarial model is built on a standard multi-layer perceptron (MLP) where the output layer only has one dimension with a sigmoid function. The  adversarial model acts as a discriminator to distinguish whether a latent code is from the prior $p_z$ (positive) or from graph encoder $\mathcal{G}(\mathbf{X, A})$ (negative). By minimizing the  cross-entropy cost for training the binary classifier, the embedding will finally be regularized and  improved during the training process. The cost can be computed as follows:
\begin{equation}
	-\frac{1}{2}\mathbb{E}_{\mathbf{z}\sim p_{z}}\text{log}  \mathcal{D}(\mathbf{Z})-\frac{1}{2}\mathbb{E}_{\mathbf{X}}\text{log}(1-\mathcal{D}(\mathcal{G}(\mathbf{X, A}))),
\end{equation}
 In our paper, we use simple Gaussian distribution as $p_z$.

\vspace{.1cm}
\noindent \textbf{Adversarial Graph Autoencoder Model. \ }
The equation for training the encoder model with Discriminator $\mathcal{D(\mathbf{Z})}$ can be written as follows: 
\begin{equation}
\small
\label{eq:objective}
	\min_{\mathcal{G}} \max_{\mathcal{D}} \mathbb{E}_{\mathbf{z}\sim p_{z} }[\text{log} \mathcal{D(\mathbf{Z})}] + \mathbb{E}_{\mathbf{x}\sim p(\mathbf{x})}[\text{log}(1 - \mathcal{D}(\mathcal{G}(\mathbf{X, A})))]
\end{equation}
where $\mathcal{G}(\mathbf{X, A})$ and $\mathcal{D(\mathbf{Z})}$ indicate the generator and discriminator explained above.

\begin {algorithm}[tpb]
\begin{small}
\caption {\small Adversarially Regularized Graph Embedding}
\label{alg:ARGE}
\begin {algorithmic}[1]
\REQUIRE
\leavevmode \\
$\mathbf{G}=\{\mathbf{V}, \mathbf{E}, \mathbf{X}\}$: a Graph with links and features;\\
$T$: the number of iterations;\\
$K$: the number of steps for iterating discriminator;\\
$d$: the dimension of the latent variable
\ENSURE
$\mathbf{Z} \in \mathbb{R}^{n \times d}$ \\
\FOR{iterator = 1,2,3, $\cdots\cdots$, $T$}
\STATE Generate latent variables matrix $\mathbf{Z}$ through Eq.(\ref{eq:latent});
\FOR{k = 1,2, $\cdots\cdots$, $K$}
\STATE Sample $m$ entities \{$\mathbf{z}^{(1)}$, \dots, $\mathbf{z}^{(m)}$\} from latent  matrix $\mathbf{Z}$
\STATE Sample $m$ entities \{$\mathbf a^{(1)}$, \dots, $\mathbf a^{(m)}$\} from the prior distribution $p_{z}$
\STATE Update the discriminator with its stochastic gradient:
\begin{equation*}
	\bigtriangledown \frac{1}{m}\sum^m_{i=1}[\text{log~}\mathcal{D}(\mathbf a^{i}) + \text{log~}(1 - \mathcal{D}( \mathbf{z}^{(i)}))]
\end{equation*}
\ENDFOR
$\mathbf{end} \ \mathbf{for}$
\STATE Update the graph autoencoder with its stochastic gradient by  Eq. (\ref{eq:autoencoder_obj0}) for ARGA or Eq. (\ref{eq:autoencoder_obj}) for ARVGA;
\ENDFOR
$\mathbf{end} \ \mathbf{for}$
\RETURN $\mathbf{Z} \in \mathbb{R}^{n \times d}$ 
\end {algorithmic}
\end{small}
\end {algorithm}
\subsection{Algorithm Explanation}
 Algorithm \ref{alg:ARGE} is our proposed framework. Given a graph $\mathbf{G}$,  the step 2 gets the latent variables matrix $\mathbf{Z}$ from the graph convolutional encoder. Then we take the same number of samples from the generated $\mathbf{Z}$ and the real data distribution $p_{z}$ in step 4 and 5 respectively, to update the discriminator with the cross-entropy cost computed in step 6. After $K$ runs of training the discriminator, the graph encoder will try to confuse the trained discriminator and update itself with generated gradient in step 7. 
 We can update Eq. (\ref{eq:autoencoder_obj0}) to train the \textbf{adversarially regularized graph autoencoder (ARGA),} or Eq. (\ref{eq:autoencoder_obj}) to train the \textbf{adversarially regularized variational graph autoencoder (ARVGA)}, respectively. Finally, we will return the graph embedding $\mathbf{Z} \in \mathbb{R}^{n \times d}$  in step 8.


\section{Experiments}

We report our results on three unsupervised graph analytic tasks: link prediction, node clustering, and graph visualization. The benchmark graph datasets used in the paper are summarized in Table 1. 
Each data set consists of scientific publications as nodes and  citation relationships as edges. The features are unique words  in each document. 
\begin{table}[htpb]
\small
  \centering
  
    \begin{tabular}{lcccccl}
    \toprule
    Data Set & \# Nodes & \# Links   & \# Content Words & \# Features\\
    \midrule
    
    Cora   &  2,708 & 5,429   & 3,880,564 & 1,433  \\
    Citeseer & 3,327 & 4,732   & 12,274,336 & 3,703  \\
    PubMed & 19,717 & 44,338  & 9,858,500   & 500 \\
    \bottomrule
    \end{tabular}%
    \caption{Real-world Graph Datasets Used in the Paper}
  \label{tab:datasets}%
\end{table}%

\subsection{Link Prediction}
\noindent\textbf{Baselines. \ } 
We compared our algorithms against state-of-the-art algorithms for the link prediction task:
\begin{itemize}
\item \textbf{DeepWalk} \cite{perozzi2014deepwalk}: is a network representation approach which encodes social relations into a continuous vector space. 
\item \textbf{Spectral Clustering} \cite{tang2011leveraging}: is an effective approach for learning social embedding.
\item \textbf{GAE} \cite{kipf2016variational}: is the most recent autoencoder-based unsupervised framework for graph data, which naturally leverages both topological and content information.
\item \textbf{VGAE} \cite{kipf2016variational}: is a variational graph autoencoder approach for graph embedding with both topological and content information. 

\item \textbf{ARGA}: Our proposed adversarially regularized autoencoder algorithm which uses graph autoencoder to learn the embedding.

\item \textbf{ARVGA}: Our proposed algorithm, which uses a \textit{variational} graph autoencoder to learn the embedding.
\end{itemize}

\noindent\textbf{Metrics. \ } We report the results in terms of AUC score  (the area under a receiver operating characteristic curve) and  average precision (AP) \cite{kipf2016variational} score. 
We conduct each experiment 10 times and report the mean values with the standard errors as the final scores. Each dataset is separated into a training, testing set and validation set. The validation set contains 5\% citation edges for hyperparameter optimization, the test set holds 10\% citation edges to verify the performance, and the rest are used for training.

\vspace{.1cm}
\noindent\textbf{Parameter Settings. }
 For the Cora and Citeseer data sets, we train all autoencoder-related models for 200 iterations and optimize them with the Adam algorithm. Both learning rate and discriminator learning rate are set as 0.001. As the PubMed data set is relatively large (around 20,000 nodes), we iterate 2,000 times for an adequate training with a 0.008 discriminator learning rate and 0.001 learning rate. We construct encoders with a 32-neuron hidden layer and a 16-neuron embedding layer for all the experiments and all the discriminators are built with two hidden layers(16-neuron, 64-neuron respectively). 
For the rest of the baselines, we retain to the settings described in the corresponding papers.





\vspace{.1cm}
\noindent\textbf{Experimental Results. \ }
The details of the experimental results on the link prediction are shown in Table 2. The results show that by incorporating an effective adversarial training module into our graph convolutional autoencoder, ARGA and ARVGA achieve outstanding performance: all AP and AUC scores are as higher as 92\% on all three data sets. Compared with all the baselines, ARGE increased the AP score from around 2.5\% compared with VGAE incorporating with node features,
11\% compared with VGAE without node features; 15.5\% and 10.6\% compared with DeepWalk and Spectral Clustering respectively on the large PubMed data set .

\begin{figure}[tpb]
\centering
\includegraphics[ width=0.49\linewidth]{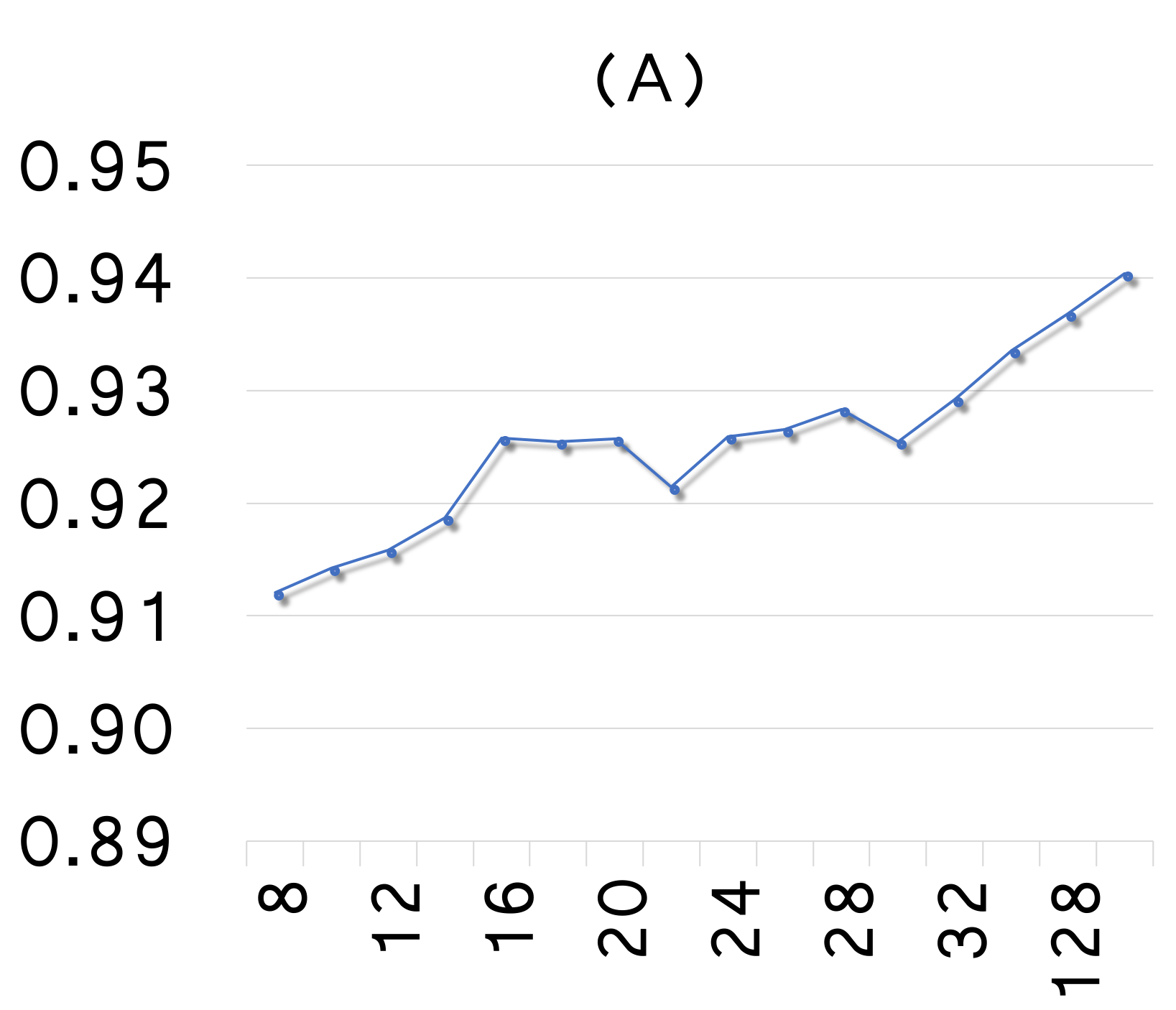}
\includegraphics[ width=0.49\linewidth]{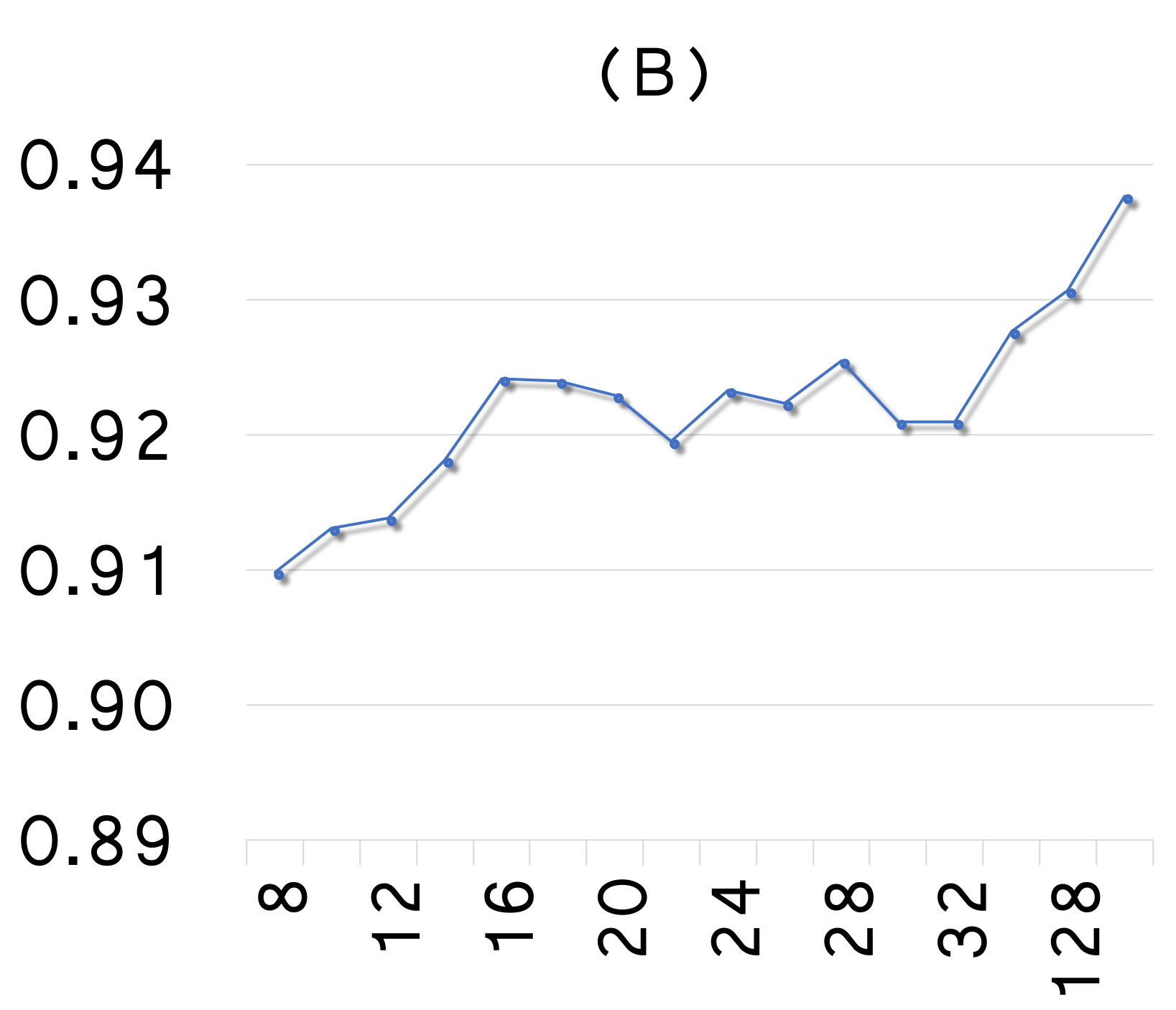}
\caption{Average performance on different dimensions of the embedding. (A) Average Precision score; (B) AUC score.} \label{fig:embedding}
\end{figure}

\noindent\textbf{Parameter Study. \ }
We vary the dimension of embedding from 8 neurons to 1024 and report the results  in Fig \ref{fig:embedding}. 

The results from both Fig \ref{fig:embedding} (A) and (B) reveal similar trends: when adding the dimension of embedding from 8-neuron to 16-neuron, the performance of embedding on link prediction steadily rises; but when we further increase the number of the neurons at the embedding layer to 32-neuron, the performance fluctuates however the results for both the AP score and the AUC score remain good.

It is worth mentioning that if we continue to set more neurons, for examples, 64-neuron, 128-neuron and 1024-neuron, the performance rises markedly. 

\begin{table*}
\small
  \centering
  
\begin{tabular}{ccccccc}
\hline
 Approaches & \multicolumn{2}{c}{Cora} & \multicolumn{2}{c}{Citeseer}& \multicolumn{2}{c}{PubMed} \\
& AUC & AP& AUC & AP& AUC & AP\\
SC &	84.6 $\pm$  0.01 &	88.5 $\pm$ 0.00 &	80.5 $\pm$  0.01 &	85.0 $\pm$  0.01 &	84.2 $\pm$  0.02 &	87.8 $\pm$  0.01\\
DW &	83.1 $\pm$  0.01 &	85.0 $\pm$  0.00 &	80.5 $\pm$ 0.02 &	83.6 $\pm$  0.01 &	84.4 $\pm$  0.00 &	84.1 $\pm$  0.00\\
      \midrule
$\text{GAE}^*$ &	84.3 $\pm$  0.02 &	88.1 $\pm$  0.01 &	78.7 $\pm$  0.02 &	84.1 $\pm$  0.02 &	82.2 $\pm$  0.01 &	87.4 $\pm$  0.00\\
$\text{VGAE}^*$ &	84.0 $\pm$  0.02 &	87.7 $\pm$  0.01 &	78.9 $\pm$  0.03 &	84.1 $\pm$  0.02 &	82.7 $\pm$  0.01 &	87.5 $\pm$  0.01\\
GAE &	91.0 $\pm$  0.02 &	92.0 $\pm$  0.03 &	89.5 $\pm$  0.04 &	89.9 $\pm$  0.05 &	96.4 $\pm$  0.00 &	96.5 $\pm$  0.00\\
VGAE &	91.4 $\pm$  0.01 &	92.6 $\pm$  0.01 &	90.8 $\pm$  0.02 &	92.0 $\pm$  0.02 &	94.4 $\pm$  0.02 &	94.7 $\pm$  0.02\\
        \midrule
\textbf{ARGE} &	92.4 $\pm$  0.003 &	\textbf{93.2 $\pm$  0.003} &	91.9 $\pm$  0.003 &	93.0$\pm$  0.003 &	\textbf{96.8 $\pm$  0.001} &	\textbf{97.1 $\pm$  0.001}\\
\textbf{ARVGE} &	\textbf{92.4 $\pm$  0.004} &	92.6 $\pm$  0.004 &	\textbf{92.4 $\pm$  0.003} &	\textbf{93.0 $\pm$  0.003} &	96.5$\pm$  0.001 &	96.8$\pm$  0.001\\
\hline
\end{tabular}
\caption{Results for Link Prediction. $\text{GAE}^*$ and $\text{VGAE}^*$ are variants of GAE, which only explore topological structure, i.e., $\mathbf X=\mathbf I$. }
  \label{tab:linkprediction}%
\end{table*}%

\begin{figure*}
\centering
\includegraphics[width=1.11in]{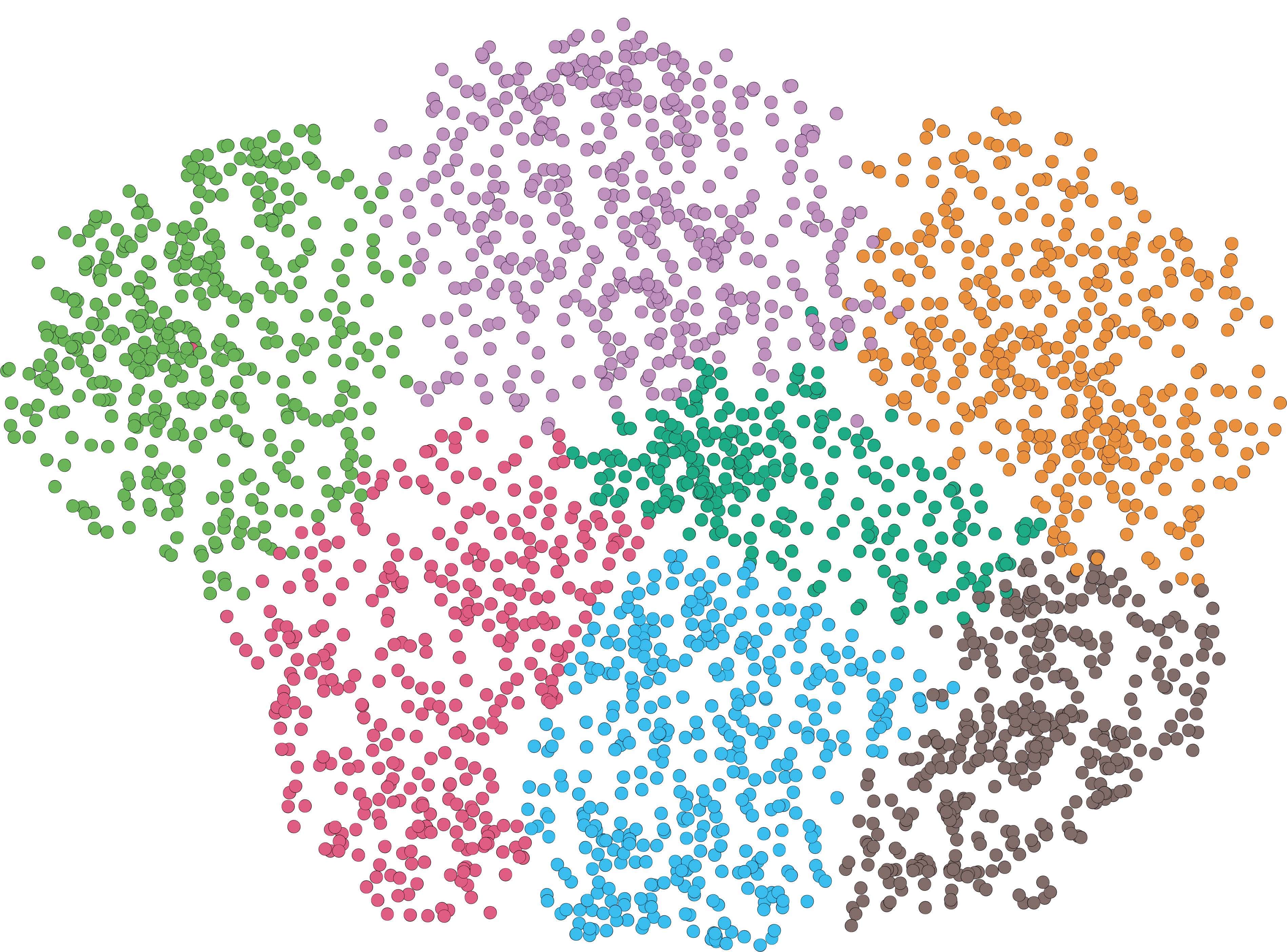}
\hspace{0.3cm}
\includegraphics[width=1.11in]{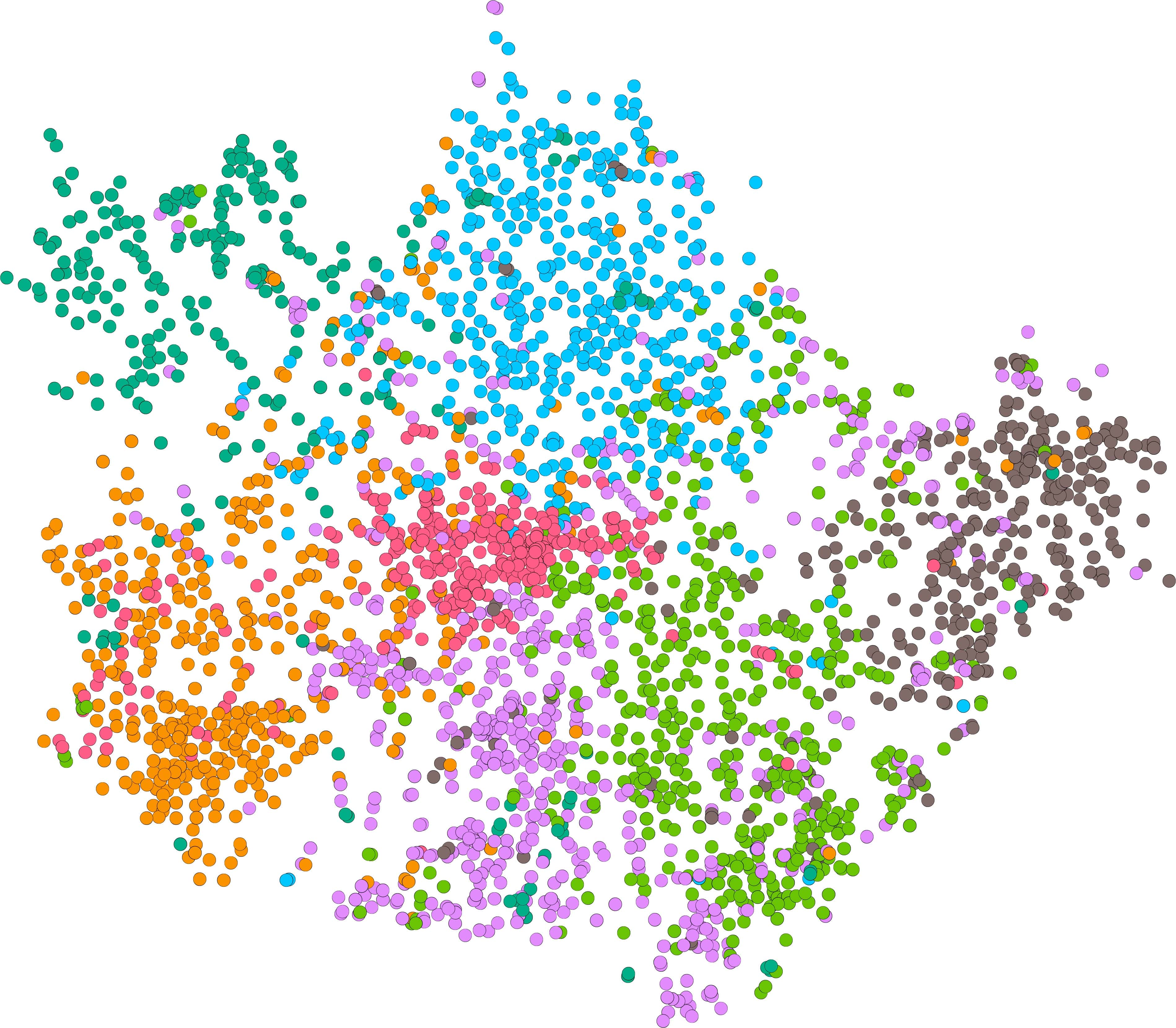}
\hspace{0.3cm}
\includegraphics[width=1.11in]{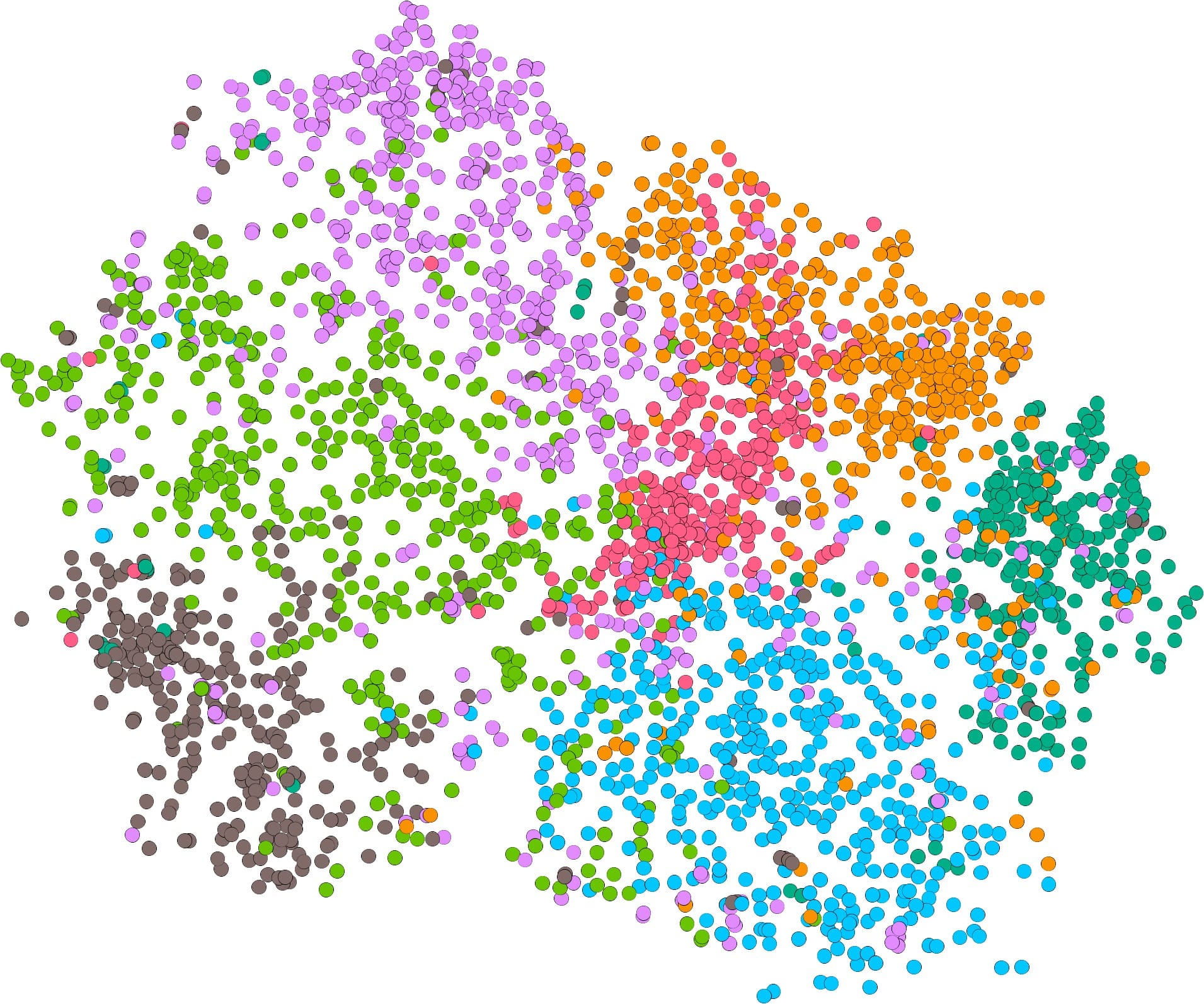}
\hspace{0.3cm}
\includegraphics[width=1.11in]{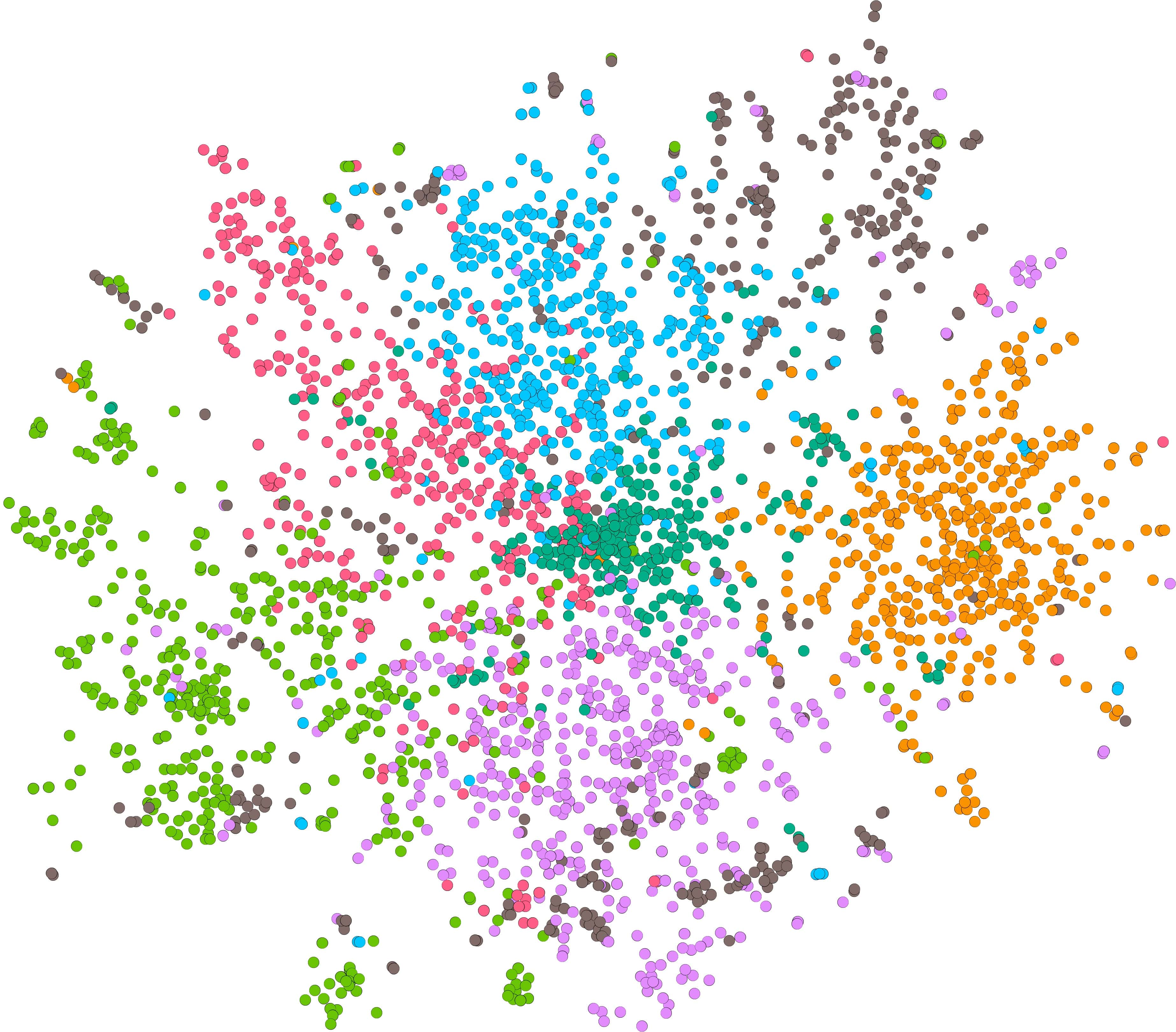}
\hspace{0.3cm}
\includegraphics[width=1.11in]{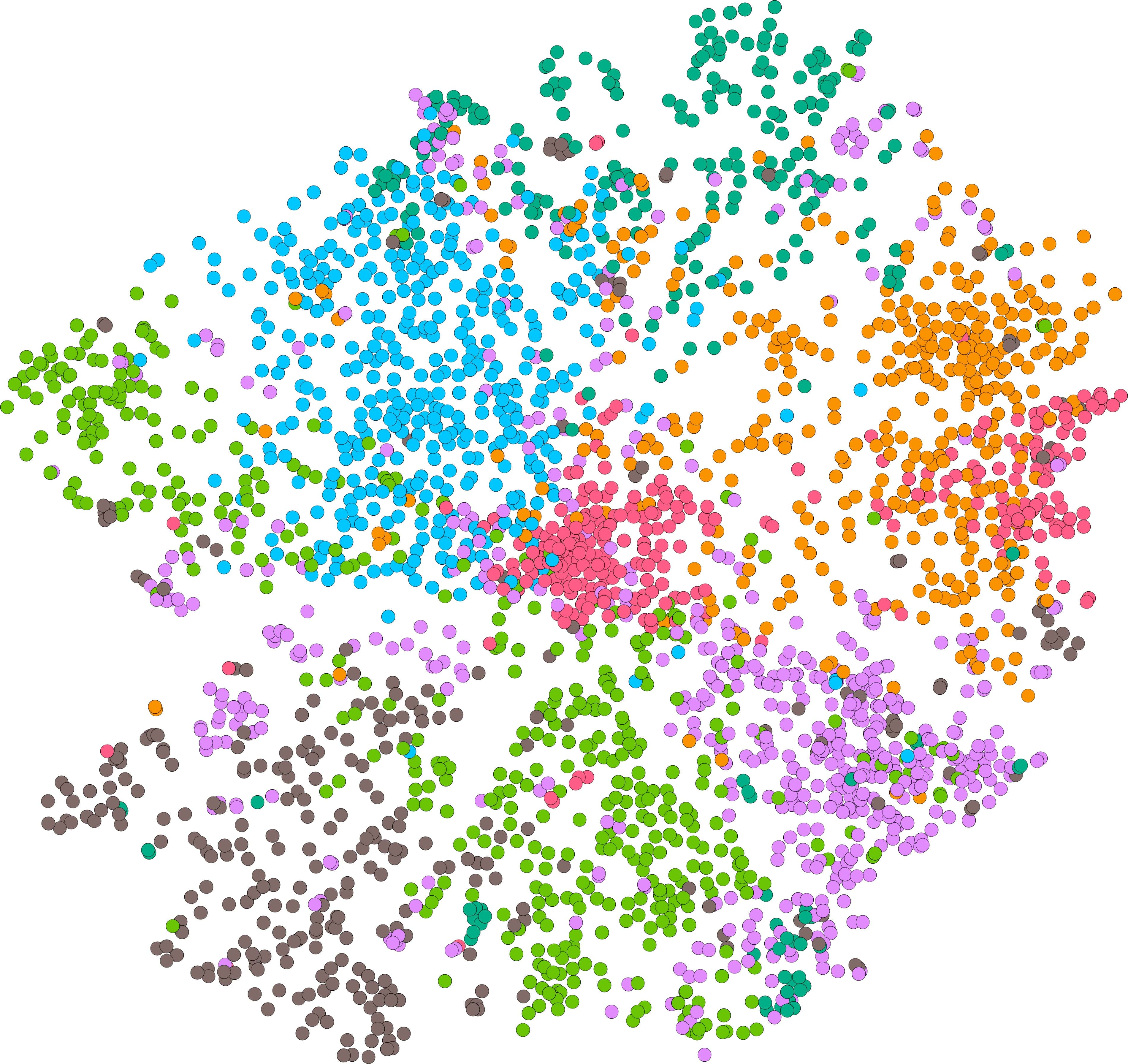}
\caption{The Cora data visualization comparison. From left to right:  embeddings from our ARGA, VGAE, GAE, DeepWalk, and Spectral Clustering. The different colors represent different groups.
} \label{fig:Visualization}
\end{figure*}

\subsection{Node Clustering}
For the node clustering task, we first learn the graph embedding, and then perform K-means clustering algorithm based on the embedding. 

\noindent\textbf{Baselines.}
We compare both embedding based approaches as well as approaches directly for graph clustering. Except for the baselines we compared for link prediction, we also include baselines which are designed for clustering:

\begin{enumerate}

\item \textbf{K-means} is a classical method and also the foundation of many clustering algorithms.
\item \textbf{Graph Encoder} \cite{tian2014learning} learns graph embedding for spectral graph clustering.
\item \textbf{DNGR} \cite{cao2016deep}  trains a stacked denoising autoencoder for graph embedding. 
\item \textbf{RTM} \cite{chang2009relational} learns the topic distributions of each document from both text and citation.
\item \textbf{RMSC} \cite{xia2014robust} employs a multi-view learning approach for graph clustering. 
\item \textbf{TADW} \cite{yang2015network} applies matrix factorization for network representation learning.
\end{enumerate}
Here the first three algorithms only exploit the graph structures, while the last three algorithms use both graph structure and node content for the graph clustering task.

\vspace{.1cm}
\noindent\textbf{Metrics. \ }
Following \cite{xia2014robust}, we employ five metrics to validate the clustering results: accuracy (Acc), normalized mutual information (NMI), precision, F-score (F1) and average rand index (ARI).

\vspace{.1cm}
\noindent\textbf{Experimental Results. \ }
The clustering results on the Cora and Citeseer data sets are given in Table 3 and Table 4. The results show that ARGA and ARVGA have achieved a dramatic improvement on all five metrics compared with all the other baselines. For instance, on Citeseer, ARGA has increased the accuracy from 6.1\% compared with K-means to 154.7\% compared with GraphEncoder; increased the F1 score from 31.9\% compared with TADW to 102.2\% compared with DeepWalk; and increased NMI from 14.8\% compared with K-means to 124.4\% compared with VGAE. 
The wide margin in the results between ARGE and GAE (and the others) has further proved the superiority of our adversarially regularized graph autoencoder.  

\begin{table}[tpb]
\small
  \centering
  
    \begin{tabular}{lccccccl}
    \toprule
    \textbf{Cora} & Acc & NMI & F1 & Precision & ARI\\
    \midrule
K-means & 0.492 & 0.321 & 0.368 & 0.369  & 0.230\\
Spectral & 0.367 & 0.127 & 0.318 & 0.193 &  0.031\\
GraphEncoder & 0.325 & 0.109 & 0.298 & 0.182& 0.006\\
DeepWalk  & 0.484 & 0.327 & 0.392 & 0.361  & 0.243\\
DNGR & 0.419 & 0.318 & 0.340 & 0.266  & 0.142\\
    \midrule
RTM & 0.440 & 0.230 & 0.307 & 0.332 &  0.169\\
RMSC & 0.407 & 0.255 & 0.331 & 0.227 &  0.090\\
TADW & 0.560 & 0.441 & 0.481 & 0.396 & 0.332\\
    \midrule
GAE & 0.596 & 0.429 & 0.595 & 0.596 & 0.347\\
VGAE & 0.609 & 0.436 & 0.609 & 0.609 & 0.346\\
    \midrule
\textbf{ARGE} & \textbf{0.640} & 0.449 & 0.619 & \textbf{0.646} & 0.352\\
\textbf{ARVGE} & 0.638 & \textbf{0.450} & \textbf{0.627} & 0.624 & \textbf{0.374}\\
        \bottomrule
    \end{tabular}%
    \caption{Clustering Results on Cora}
  \label{tab:clustering_cora}%
\end{table}%





\subsection{Graph Visualization}
We  visualize the Cora data in a two-dimensional space by applying the t-SNE algorithm \cite{van2014accelerating} on the learned embedding. The results  in Fig 3 validate that by applying adversarial training to the graph data, we can obtained a more meaningful layout of the graph data.
\begin{table}[tpb]
\small
  \centering
  
    \begin{tabular}{lccccccl}
    \toprule
\textbf{Citeseer} & Acc & NMI & F1 & Precision & ARI\\
\midrule K-means & 0.540 & 0.305 & 0.409 & 0.405 &0.279\\Spectral & 0.239 & 0.056 & 0.299 & 0.179 &  0.010\\
GraphEncoder & 0.225 & 0.033 & 0.301 & 0.179 & 0.010\\DeepWalk & 0.337 & 0.088 & 0.270 & 0.248 & 0.092\\DNGR & 0.326 & 0.180 & 0.300 & 0.200 & 0.044\\
\midrule 
RTM  & 0.451 & 0.239 & 0.342 & 0.349 &  0.203\\
RMSC & 0.295 & 0.139 & 0.320 & 0.204 &  0.049\\
TADW & 0.455 & 0.291 & 0.414 & 0.312 & 0.228\\
\midrule
GAE & 0.408 & 0.176 & 0.372 & 0.418 & 0.124\\
VGAE & 0.344 & 0.156 & 0.308 & 0.349 & 0.093\\
\midrule
\textbf{ARGE} & \textbf{0.573} & \textbf{0.350} & \textbf{0.546} & \textbf{0.573} & \textbf{0.341}\\\textbf{ARVGE} & 0.544 & 0.261 & 0.529 & 0.549 & 0.245\\
        \bottomrule
    \end{tabular}%
    \caption{Clustering Results on Citeseer}
  \label{tab:cluatering_citeseer}%
\end{table}%



\section{Conclusion}
In this paper, we proposed a novel  adversarial graph embedding framework  for graph data. We argue that most existing graph embedding algorithms are unregularized methods that ignore the   data distributions of the latent representation and  suffer from inferior embedding  in real-world graph data. We proposed an adversarial training scheme to \textit{regularize} the latent codes and enforce the latent codes to match a prior distribution. The adversarial module is jointly learned with a graph convolutional autoencoder to produce a robust representation.  
Experiment results demonstrated that our algorithms ARGA and ARVGA outperform baselines in link prediction, node clustering, and graph visualization tasks.

\section*{Acknowledgements}
This research was funded by the Australian Government through the Australian Research Council (ARC) under grants 1) LP160100630 partnership with Australia Government Department of Health and 2) LP150100671 partnership with Australia Research Alliance for Children and Youth (ARACY) and Global Business College Australia (GBCA). 
We  acknowledge the support of NVIDIA Corporation and MakeMagic Australia with the donation of GPU used for this research.


\bibliographystyle{named}
\bibliography{ijcai18}

\begin{thebibliography}{}

\bibitem[\protect\citeauthoryear{Cai \bgroup \em et al.\egroup
  }{2017}]{cai2017comprehensive}
H.~Cai, V.~W. Zheng, and K.~C-C Chang.
\newblock A comprehensive survey of graph embedding: Problems, techniques and
  applications.
\newblock {\em arXiv preprint arXiv:1709.07604}, 2017.

\bibitem[\protect\citeauthoryear{Cao \bgroup \em et al.\egroup
  }{2015}]{cao2015grarep}
S.~Cao, W.~Lu, and Q.~Xu.
\newblock Grarep: Learning graph representations with global structural
  information.
\newblock In {\em CIKM}, pages 891--900. ACM, 2015.

\bibitem[\protect\citeauthoryear{Cao \bgroup \em et al.\egroup
  }{2016}]{cao2016deep}
S.~Cao, W.~Lu, and Q.~Xu.
\newblock Deep neural networks for learning graph representations.
\newblock In {\em AAAI}, pages 1145--1152, 2016.

\bibitem[\protect\citeauthoryear{Chang and Blei}{2009}]{chang2009relational}
J.~Chang and D.~Blei.
\newblock Relational topic models for document networks.
\newblock In {\em Artificial Intelligence and Statistics}, pages 81--88, 2009.

\bibitem[\protect\citeauthoryear{Cui \bgroup \em et al.\egroup
  }{2017}]{cui2017survey}
P.~Cui, X.~Wang, J.~Pei, and et.al.
\newblock A survey on network embedding.
\newblock {\em arXiv preprint arXiv:1711.08752}, 2017.

\bibitem[\protect\citeauthoryear{Dai \bgroup \em et al.\egroup
  }{2017}]{dai2017adversarial}
Q.~Dai, Q.~Li, J.~Tang, and et.al.
\newblock Adversarial network embedding.
\newblock {\em arXiv preprint arXiv:1711.07838}, 2017.

\bibitem[\protect\citeauthoryear{Donahue \bgroup \em et al.\egroup
  }{2016}]{donahue2016adversarial}
J.~Donahue, P.~Kr{\"a}henb{\"u}hl, and T.~Darrell.
\newblock Adversarial feature learning.
\newblock {\em arXiv preprint arXiv:1605.09782}, 2016.

\bibitem[\protect\citeauthoryear{Goodfellow \bgroup \em et al.\egroup
  }{2014}]{goodfellow2014generative}
I.~Goodfellow, J.~Pouget-Abadie, M.~Mirza, and et.al.
\newblock Generative adversarial nets.
\newblock In {\em NIPS}, pages 2672--2680, 2014.

\bibitem[\protect\citeauthoryear{Goyal and Ferrara}{2017}]{goyal2017graph}
P.~Goyal and E.~Ferrara.
\newblock Graph embedding techniques, applications, and performance: A survey.
\newblock {\em arXiv preprint arXiv:1705.02801}, 2017.

\bibitem[\protect\citeauthoryear{Grover and
  Leskovec}{2016}]{grover2016node2vec}
A.~Grover and J.~Leskovec.
\newblock node2vec: Scalable feature learning for networks.
\newblock In {\em SIGKDD}, pages 855--864. ACM, 2016.

\bibitem[\protect\citeauthoryear{Kipf and Welling}{2016a}]{kipf2016semi}
T.~N Kipf and M.~Welling.
\newblock Semi-supervised classification with graph convolutional networks.
\newblock {\em arXiv preprint arXiv:1609.02907}, 2016.

\bibitem[\protect\citeauthoryear{Kipf and Welling}{2016b}]{kipf2016variational}
T.~N Kipf and M.~Welling.
\newblock Variational graph auto-encoders.
\newblock {\em NIPS}, 2016.

\bibitem[\protect\citeauthoryear{Maaten}{2014}]{van2014accelerating}
L.~V.~D. Maaten.
\newblock Accelerating t-sne using tree-based algorithms.
\newblock {\em JMLR}, 15(1):3221--3245, 2014.

\bibitem[\protect\citeauthoryear{Makhzani \bgroup \em et al.\egroup
  }{2015}]{makhzani2015adversarial}
A.~Makhzani, J.~Shlens, N.~Jaitly, and et.al.
\newblock Adversarial autoencoders.
\newblock {\em arXiv preprint arXiv:1511.05644}, 2015.

\bibitem[\protect\citeauthoryear{Ou \bgroup \em et al.\egroup
  }{2016}]{ou2016asymmetric}
M.~Ou, P.~Cui, J.~Pei, and et.al.
\newblock Asymmetric transitivity preserving graph embedding.
\newblock In {\em KDD}, pages 1105--1114, 2016.

\bibitem[\protect\citeauthoryear{Pan \bgroup \em et al.\egroup
  }{2016a}]{pan2016joint}
S.~Pan, J.~Wu, X.~Zhu, and et.al.
\newblock Joint structure feature exploration and regularization for multi-task
  graph classification.
\newblock {\em TKDE}, 28(3):715--728, 2016.

\bibitem[\protect\citeauthoryear{Pan \bgroup \em et al.\egroup
  }{2016b}]{pan2016tri}
S.~Pan, J.~Wu, X.~Zhu, and et.al.
\newblock Tri-party deep network representation.
\newblock In {\em IJCAI}, pages 1895--1901, 2016.

\bibitem[\protect\citeauthoryear{Perozzi \bgroup \em et al.\egroup
  }{2014}]{perozzi2014deepwalk}
B.~Perozzi, R.~Al-Rfou, and S.~Skiena.
\newblock Deepwalk: Online learning of social representations.
\newblock In {\em SIGKDD}, pages 701--710. ACM, 2014.

\bibitem[\protect\citeauthoryear{Qiu \bgroup \em et al.\egroup
  }{2017}]{qiu2017network}
J.~Qiu, Y.~Dong, H.~Ma, and et.al.
\newblock Network embedding as matrix factorization: Unifyingdeepwalk, line,
  pte, and node2vec.
\newblock {\em arXiv preprint arXiv:1710.02971}, 2017.

\bibitem[\protect\citeauthoryear{Radford \bgroup \em et al.\egroup
  }{2015}]{radford2015unsupervised}
A.~Radford, L.~Metz, and S.~Chintala.
\newblock Unsupervised representation learning with deep convolutional
  generative adversarial networks.
\newblock {\em arXiv preprint arXiv:1511.06434}, 2015.

\bibitem[\protect\citeauthoryear{Tang and Liu}{2011}]{tang2011leveraging}
L.~Tang and H.~Liu.
\newblock Leveraging social media networks for classification.
\newblock {\em DMKD}, 23(3):447--478, 2011.

\bibitem[\protect\citeauthoryear{Tang \bgroup \em et al.\egroup
  }{2015}]{Tang2015}
J.~Tang, M.~Qu, M.~Wang, and et.al.
\newblock {LINE:} large-scale information network embedding.
\newblock In {\em WWW}, pages 1067--1077, 2015.

\bibitem[\protect\citeauthoryear{Tian \bgroup \em et al.\egroup
  }{2014}]{tian2014learning}
F.~Tian, B.~Gao, Q.~Cui, and et.al.
\newblock Learning deep representations for graph clustering.
\newblock In {\em AAAI}, pages 1293--1299, 2014.

\bibitem[\protect\citeauthoryear{Wang \bgroup \em et al.\egroup
  }{2016}]{wang2016structural}
D.~Wang, P.~Cui, and W.~Zhu.
\newblock Structural deep network embedding.
\newblock In {\em SIGKDD}, pages 1225--1234. ACM, 2016.

\bibitem[\protect\citeauthoryear{Wang \bgroup \em et al.\egroup
  }{2017a}]{wang2017mgae}
C.~Wang, S.~Pan, G.~Long, and et.al.
\newblock Mgae: Marginalized graph autoencoder for graph clustering.
\newblock In {\em CIKM}, pages 889--898. ACM, 2017.

\bibitem[\protect\citeauthoryear{Wang \bgroup \em et al.\egroup
  }{2017b}]{wang2017community}
X.~Wang, P.~Cui, J.~Wang, and et.al.
\newblock Community preserving network embedding.
\newblock In {\em AAAI}, pages 203--209, 2017.

\bibitem[\protect\citeauthoryear{Wang \bgroup \em et al.\egroup
  }{2017c}]{wang2017predictive}
Z.~Wang, C.~Chen, and W.~Li.
\newblock Predictive network representation learning for link prediction.
\newblock In {\em SIGIR}, pages 969--972. ACM, 2017.

\bibitem[\protect\citeauthoryear{Xia \bgroup \em et al.\egroup
  }{2014}]{xia2014robust}
R.~Xia, Y.~Pan, L.~Du, and et.al.
\newblock Robust multi-view spectral clustering via low-rank and sparse
  decomposition.
\newblock In {\em AAAI}, pages 2149--2155, 2014.

\bibitem[\protect\citeauthoryear{Yang \bgroup \em et al.\egroup
  }{2015}]{yang2015network}
C.~Yang, Z.~Liu, D.~Zhao, and et.al.
\newblock Network representation learning with rich text information.
\newblock In {\em IJCAI}, pages 2111--2117, 2015.

\bibitem[\protect\citeauthoryear{Zhang \bgroup \em et al.\egroup
  }{2017a}]{zhang2017network}
D.~Zhang, J.~Yin, X.~Zhu, and et.al.
\newblock Network representation learning: A survey.
\newblock {\em arXiv preprint arXiv:1801.05852}, 2017.

\bibitem[\protect\citeauthoryear{Zhang \bgroup \em et al.\egroup
  }{2017b}]{zhang2017user}
D.~Zhang, J.~Yin, X.~Zhu, and et.al.
\newblock User profile preserving social network embedding.
\newblock In {\em IJCAI}, pages 3378--3384, 2017.

\end{thebibliography}

\end{document}